# TÜRKÇE ADRES AYRIŞTIRMA İÇİN ÖNCEDEN EĞİTİLMİŞ DİL MODELLERİNİN KARŞILAŞTIRILMASI
# COMPARISON OF PRE-TRAINED LANGUAGE MODELS FOR TURKISH ADDRESS PARSING

Muhammed Cihat Ünal
Huawei Turkey R&D Center, Istanbul
Betül Aygün
Huawei Turkiye R&D Center, Istanbul
Aydın Gerek
Huawei Turkiye R&D Center, Istanbul

**Özet**: Büyük veri kümesi üzerinde eğitilen BERT ve varyantları gibi transformer tabanlı önceden eğitilmiş dil modelleri, doğal dil işleme (NLP) görevleri için büyük bir başarı göstermektedir. Bu alanda yapılan akademik çalışmaların çoğu İngilizce diline dayanmaktadır; ancak çok dilli ve dile özgü çalışmalar da giderek artmaktadır. Ayrıca, çeşitli araştırmalar, dile özgü bazı modellerin çeşitli görevlerde çok dilli modellerden daha iyi performans gösterdiğini iddia etmektedir. Bu nedenle araştırmacılar, modelleri kendi durumları için özel olarak eğitim veya ince ayar yapma eğilimindedir. Bu yazıda, Türkiye harita verilerine odaklanıyoruz ve BERT, Distill-BERT, ELECTRA ve RoBERTa'nın hem çok dilli hem de Türkçe tabanlılarını kapsamlı bir şekilde değerlendiriyoruz. Ayrıca, tek katmanlı ince ayarın standart yaklaşımına ek olarak BERT'nin ince ayarı için bir Çok Katmanlı Algılayıcı (MLP) öneriyoruz. Veri kümesi için, nispeten yüksek kalitede orta boyutlu Adres Ayrıştırma veri seti kullanıldı. Bu veri kümesi üzerinde yapılan deneyler, MLP ince ayarlı Türkçe'ye özgü modellerin, çok dilli ince ayarlı modellere kıyasla biraz daha iyi sonuçlar verdiğini göstermektedir. Ayrıca, adres belirteçlerinin temsillerinin görselleştirilmesi, çeşitli adresleri sınıflandırmak için BERT varyantlarının etkinliğini de gösterir.
**Anahtar Kelimeler**: Bert, Distill-Bert, M-Bert, XLM-Roberta, MLP, Harita Verisi

**Abstract**: Transformer based pre-trained models such as BERT and its variants, which are trained on large corpora, have demonstrated tremendous success for natural language processing (NLP) tasks. Most of academic works are based on the English language; however, the number of multilingual and language specific studies increase steadily. Furthermore, several studies claimed that language specific models outperform multilingual models in various tasks. Therefore, the community tends to train or fine-tune the models for the language of their case study, specifically. In this paper, we focus on Turkish maps data and thoroughly evaluate both multilingual and Turkish based BERT, DistilBERT, ELECTRA and RoBERTa. Besides, we also propose a Multi-Layer Perceptron (MLP) for fine-tuning BERT in addition to the standard approach of one-layer fine-tuning. For the dataset, a mid-sized Address Parsing corpus taken with a relatively high quality is constructed. Conducted experiments on this dataset indicate that Turkish language specific models with MLP fine-tuning yields slightly better results when compared to the multilingual fine-tuned models. Moreover, visualization of address tokens' representations further indicates the effectiveness of BERT variants for classifying a variety of addresses.

**Introduction**

Addresses are key resources for people and organizations in the modern world. Addresses include valuable information for a variety of software systems such as estimation systems, recommendation systems, e-commerce software systems and maps applications (ex. petal maps, google maps). Success of these applications dependent on the information that are extracted from Addresses data. However, extracting relevant information such as country, city, floor number or even point of interest (POI) from addresses is not an easy task most of the time, especially for the addresses entered by the end users due to the high number of possibilities of each token to take. Address parsing is an important task that attempts to find a label for each entity in an address, such as a city, country, or neighborhood to increase the quality of the software, and it is necessary to do it properly for aforementioned software systems. Although in the literature, there is a huge number of studies for address parsing in the English language, for the Turkish based addresses, there is no significant study.

In this paper, we focus on address parsing for Turkish language from a token classification approach. Token classification is one of the most common NLP tasks. It can be used for various purposes such as named entity recognition (NER), part-of-speech (POS) tagging, slot filing and address parsing. There are lots of ways to achieve these tasks. One way is using statistical approaches, where conditional random fields (CRF) [1] been proven to be an effective framework, by taking discrete features as the representation of input sequence [2]. Thanks to developments in deep learning, neural token classification models have achieved state-of-the-art results for many tasks.

The goal of our experiments is providing probable solutions for Turkish address parsing by language-specific and multilingual BERT models. When it comes to dealing with such an NLP problem, BERT [3] and its variants (such as M-BERT [4], DistilBERT [5], RoBERTa [6]) are the models that we can count on in terms of their performance and success. In our work, we mostly used transformer based pre-trained model for Turkish, but also, we used M-BERT. M-BERT is a multilingual BERT model trained on pooled data from 104 languages. While M-BERT has been shown to have a tremendous ability to generalize across languages [4] several studies have also concluded that language-specific BERT models, can noticeably outperform M-BERT. FinBERT [7] and CamemBERT [8] are such examples. Therefore, we added Turkish specific BERT models besides M-BERT to our trials.

These models have been pre-trained on more than 2M words, and their multilayered, complex architecture makes them powerful models to use. Therefore, our solutions are based on fine-tuning pre-trained BERT [3] models, and customization of some of these pre-trained models. We've added multiple linear layers with an activation function, dropouts and batch normalization to the top of the BERT [3] architecture in terms of customization. In standard fine-tuning approach, a linear classifier layer is composed with the outputs of BERT like models. One architectural choice we have experimented with was using a two-layer MLP instead of a linear classifier. The reason of why we've done customization to the top of BERT model is: emphasizing the customization of pre-trained models, even as strong as BERT, can surpass the success of the fine-tuning. We used random search technique by more than forty trials for the hyper-parameter optimization to decide best parameters to be used in training for obtaining more robust results.

Finally, we've compared seven pre-trained models with additional MLP architecture by measuring the accuracy (for per token and sample), recall, precision, and thus F1 score metrics of the prediction scores. Since we have 25 different labels in our case (e.g., district, village and country), we will evaluate these success measures of models for 25 different labels. Confusion matrix will be used for aforementioned metrics' calculation. Our main contribution of this work is to show the existing prominent pre-trained models' performances in Turkish address parsing tasks. Besides, we also claim that adding a more complex architecture for fine-tuning also improves the accuracy of the classification of tokens.

**Literature Review**

Pre-trained models have become an indispensable part of developing algorithms and building models for not only natural language understanding, but also various machine learning tasks. They have been trained on large corpora. This allows them to provide promising results for many tasks in machine learning, even though some additional layers or different architecture may be required to get more reasonable results for some cases. Although all pre-trained models are claimed that they yields better results, the success of these models depends on the nature of the data, number of labels and task and the value of the hyper-parameters. Since there are more than one pre-trained models, and some of them are trained for specific tasks or languages (e.g. ELECTRA, M-BERT), you need to decide which one you should pick in your cases.

During this study, we focus on the Turkish language of the maps data with multiple labels. In the literature, there is a huge number of studies conducted on different languages and on different tasks that compare pre-trained models [9] [10] . The domain of the tasks conducted by the mainly of these studies extends from sentiment analysis to financial domain in a wide range areas. [11] [12] [13]  However, very few experiments have been conducted on maps data of the Turkish language. Turkish based languages are mostly studied on different subjects such as classification of medical text documents, sentiment analysis, and offensive language detection [14] [15] [16] [17] [18] [19].

**Methodology**

*Model Selection*

We would like to estimate correctly the type of words or word groups by using unstructured address queries of users' as input. As discussed in the introduction, this study compares the success of pre-trained BERT model and its variants (ELECTRA [20], RoBERTa [6], DistilBERT [5]) besides with additional MLP architecture for fine tuning. In the study, we compare these successful pre-trained language models in their capacity for the address parsing tasks in the Turkish language. This comparison allows us to demonstrate which model has the highest efficiency.

The models which are used for fine-tuning in this study are:

- bert-base-multilingual-uncased
- bert-base-multilingual-cased
- distilbert-base-multilingual-cased
- dbmdz/distilbert-base-turkish-cased
- savasy/bert-base-turkish-ner-cased
- mys/electra-base-turkish-cased-ner
- akdeniz27/xlm-roberta-base-turkish-ner

*Hyperparameter Optimization*

As we discussed in the introduction, we need to do hyperparameter optimization for each model to obtain best results for robust comparison. We used Optuna library for this purpose. Optuna is a python library that enables to tune our machine learning models automatically [21]. It is applicable to ubiquitous machine learning frameworks such as TensorFlow, PyTorch, LightGBM and Sklearn. We use PyTorch for the experiments. Optuna will try to find the best values for the parameters of "learning rate", "weight decay", "batch size" and "optimizer" by random search with 40 trials. The range of all hyper-parameters is represented in the Table 1. Learning rate will be a random floating number that will be selected in range between 5e-5 and 1e-2 for each iteration. For the optimizer, AdamW [22] [23] and SGD [24] is selected. For the batch size, $3^{rd}$, $4^{th}$, $5^{th}$ and $6^{th}$ powers of 2 are tried in the Optuna.

Table 1. Optional Parameters' values used in OPTUNA

| *Parameters* | *Options* |
|---|---|
| *Learning Rate* | [5e-5, 1e-2] |
| *Batch Size* | {8, 16, 32, 64} |
| *Optimizer* | {AdamW, RMSprop, SGD} |
| *Weight Decay* | [1e-3, 1e-2, 1e-4] |

*Training Details*

We convert all data to lowercase because character "i" for Turkish is different than English one. English character "i" resides with some special character next to it. In consequence, we encounter with some problems in tokenization part. Surely, tokenizer doesn't ignore this special character and cannot manage tokenization successfully. Eventually, we convert English character "i" to Turkish character "i" along with converting all other characters to lowercase for making all characters coherent with Turkish. Thus, we avoid having same kind of problem with any other characters.

We compare 14 models in total (seven models plus their MLP added architecture). The architecture of the MLP is two hidden linear layer with one dropout layer with 0.4 probability. Hyper-parameters are tuned on the validation set. According to the maximum length of the input query, the maximum padding length becomes 256. Linear scheduler is used as a scheduler and maximum number of epochs is 10 for every iteration. Since we use early stopping based on validation loss to terminate unpromising iterations, number of epochs can be vary for each iteration. Thus, wasting time and resources will be avoided.

**Experiments and Results**

*Dataset Description*

The unlabeled dataset is collected by the address queries from Petal Maps [25] application entered by real end users. The size of the dataset is 1248 and divided into three parts: train, validation and test. Train data is %70 of the whole data, and validation and test datasets cover half of the remaining data (%30). Since we are dealing with slot filling/address parsing task, we put the data in CoNLL format. Unlabeled data set is annotated by the annotators. In total there are

14 different labels in our data, with regards to the IOB format [26] the total number of different labels become 25. In detail, we annotate each word as being either other ("O"), inside an entity ("I-COUNTRY") or at the beginning of entity ("B-COUNTRY"). It is important to use this format because we want to make sure that models can understand the differences between addresses. Imagine that your text is "nike store hagia sofia", where "nike store" and "hagia sofia" are two different POI. If you want to make sure the model understands this difference, you should use IOB to check that the result is "B-POI I-POI B-POI I-POI". Figure 1 illustrates the frequency of the labels. It indicates that the labels are imbalanced. POI has the highest frequency, on the other hand, the number of "door" has the lowest frequency.

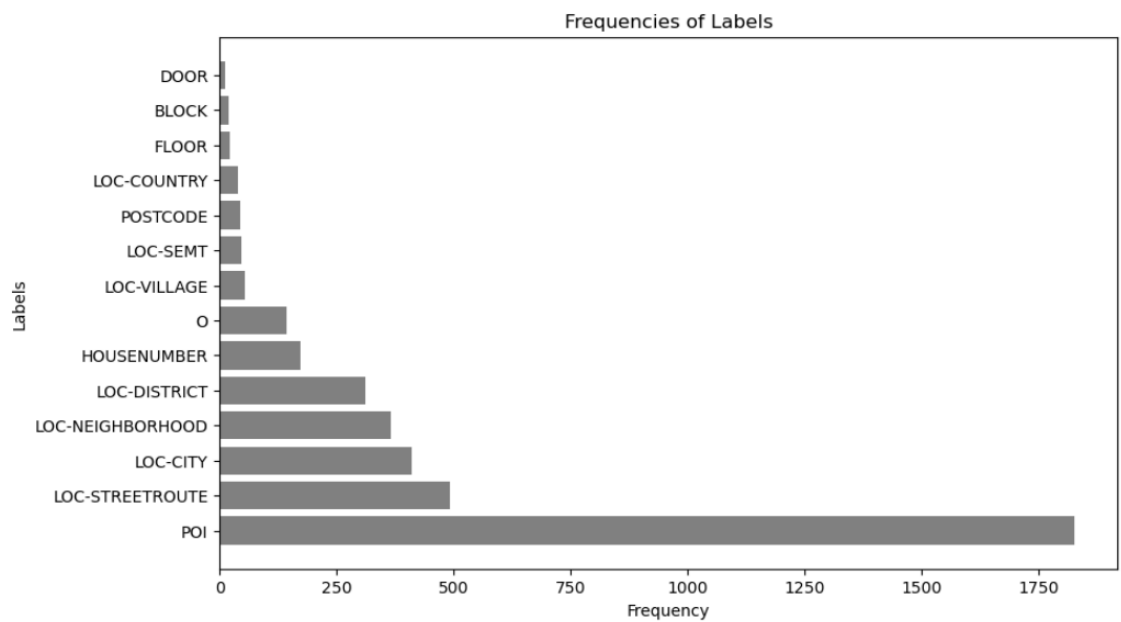

Figure 1. Frequencies of the labels in the dataset

*Parameter Values for the Models*

The several options for the parameters used in the Optuna are represented in the previous Chapter. According to them, hyper-parameter optimization values evaluated by Optuna for each model are shown in Table 2.

Table 2. Hyperparameter optimization values for each model by Optuna

| Models | Hyper-parameter values | | | |
|---|---|---|---|---|
| | *Learning Rate* | *Optimizer* | *Weight Decay* | *Batch Size* |
| bert-base-multilingual-uncased_finetuned | 6.114e-05 | AdamW | 0.0001 | 32 |
| bert-base-multilingual-uncased_MLP | 1.761e-05 | AdamW | 0.01 | 32 |
| bert-base-multilingual-cased_finetuned | 3.359e-05 | AdamW | 0.001 | 32 |
| bert-base-multilingual-cased_MLP | 7.948e-05 | AdamW | 0.01 | 16 |
| distilbert-base-multilingual-cased_finetuned | 5.703e-05 | AdamW | 0.01 | 32 |
| distilbert-base-multilingual-cased_MLP | 7.948e-05 | AdamW | 0.01 | 16 |
| dbmdz/distilbert-base-turkish-cased_finetuned | 6.861e-05 | AdamW | 0.0001 | 32 |
| dbmdz/distilbert-base-turkish-cased_MLP | 8.867e-05 | RMSprop | 0.001 | 64 |
| savasy/bert-base-turkish-ner-cased_finetuned | 2.546e-05 | AdamW | 0.001 | 32 |
| savasy/bert-base-turkish-ner-cased_MLP | 3.069e-05 | RMSprop | 0.001 | 32 |
| mys/electra-base-turkish-cased-ner_finetuned | 6.861e-05 | AdamW | 0.0001 | 32 |
| mys/electra-base-turkish-cased-ner_MLP | 3.217e-05 | AdamW | 0.01 | 32 |
| akdeniz27/xlm-roberta-base-turkish-ner_finetuned | 1.455e-05 | AdamW | 0.0001 | 8 |
| akdeniz27/xlm-roberta-base-turkish-ner_MLP | 5.599e-05 | AdamW | 0.0001 | 64 |

*Performance Metrics*

The performance of the Bert models is measured in following ways:

1. Sample Wise Accuracy: This is the percentage of samples (queries) correctly classified by the models. For example, for an address, if all tokens are correctly predicted, then it is counted as correct sample. So, if 8 out of 10 samples are correctly predicted, the accuracy is 80 percent. This is depicted as in the following equation.

$$Accuracy = \frac{100 * N}{S}$$

Where N represents the number of samples correctly predicted and S represents the total number of samples.

2. Token Wise Accuracy: This is the percentage of tokens correctly classified by the models. We count token count and correctly predicted token count for each address sample. At the end, ratio between correctly predicted token and token count gives the accuracy. To illustrate, assumingly two different addresses have 10 token in total. If 5 out of 10 tokens are correctly classified, the accuracy is 50 percent. This is depicted as in the following equation.

$$Accuracy = \frac{100 * C}{T}$$

Where C represents the number of tokens correctly classified and T represents the total number of tokens.

3. Macro Precision: Since there are 25 unique labels for tokens, precision has to be calculated for each label first. Thus, 25 different precision scores are obtained. Finally, we take the average of these scores to measure models' precision. The final score is called "macro precision".

$$Precision = \frac{TP}{TP + FP}$$

Number of True Positives (TP) divided by the Total Number of True Positives (TP) and False Positives (FP).

$$Macro\ Precision = \frac{\sum_{i=1}^{T} p(i)}{T}$$

Where T represents the total label count and p(i) represents the precision score of the pertinent label.

4. Macro Recall: Since there are 25 unique labels for tokens, recall has to be calculated for each label first. Next, 25 different recall scores are obtained. At the end, we take the average of these scores to measure models' recall. The final score is called "macro recall".

$$Recall = \frac{TP}{TP + FN}$$

Number of True Positives (TP) divided by the Total Number of True Positives (TP) and False Negatives (FN).

$$Macro\ Recall = \frac{\sum_{i=1}^{T} r(i)}{T}$$

Where T represents the total label count and r(i) represents the recall score of the pertinent label.

5. Macro F1: Since there are 25 unique labels for tokens, F1 score has to be calculated for each label first. Next, 25 different F1 scores are obtained. At the end, we take the average of these scores to measure models' F1 scores. The final score is called "macro F1 score".

$$F1\ Score = 2 * \frac{Precision * Recall}{Precision + Recall}$$

$$Macro\ F1\ Score = \frac{\sum_{i=1}^{T} f(i)}{T}$$

Where T represents the total label count and f(i) represents the F1 score of the pertinent label.

*Results and Analysis*

After the hyperparameter optimization, metrics are evaluated for all models on the test set. Experiments are conducted under 1000 number of epochs. The results of the all metrics formulated are shown in the Table 2. We have evaluated accuracy for per sample and per token. Since this is a token classification task, it is not possible to calculate precision and recall for per sample. Therefore, we calculate these metrics for each token and use macro calculations for these metrics.

According to the Table 3, fine-tuned models are mostly the successful ones in terms of per sample accuracy when we compare them with their additional MLP version. However, *distilbert-base-multilingual-cased* model with additional MLP outperforms its fine-tuned model.

In terms of per token accuracy, additional MLP version of some models are more successful than their fine-tuning; such as *bert-base-multilingual-cased*, *distil-bert-multilingual-cased*, *dbmdz/distilbert-base-turkish-cased* and *mys/electra-base-turkish-cased-ner*. Moreover, even though fine-tuned version of some models have higher per sample accuracy, they result in lower precision, recall and F1 score against their additional MLP version. For example, *dbmdz/distilbert-base-turkish-cased* model's fine-tuned model has higher per sample accuracy, but its fine-tuned version has lower precision, recall and F1 score than its additional MLP version. Besides, we can also conclude that models specific for the Turkish language yield better results than multilingual models.

Table 3. Evaluation Results of Models

| Models | Success Metrics | | | | |
| --- | --- | --- | --- | --- | --- |
| | *Precision (macro)* | *Recall (macro)* | *F1 (macro)* | *Accuracy (Per Sample)* | *Accuracy (Per Token)* |
| bert-base-multilingual-uncased_finetuned | 0.419 | 0.460 | 0.429 | 0.622 | 0.743 |
| bert-base-multilingual-uncased_MLP | 0.370 | 0.398 | 0.378 | 0.585 | 0.726 |
| bert-base-multilingual-cased_finetuned | 0.385 | 0.412 | 0.389 | 0.585 | 0.709 |
| bert-base-multilingual-cased_MLP | 0.339 | 0.396 | 0.350 | 0.569 | 0.72 |
| distilbert-base-multilingual-cased_finetuned | 0.425 | 0.410 | 0.398 | 0.564 | 0.694 |
| distilbert-base-multilingual-cased_MLP | 0.475 | 0.471 | 0.452 | 0.559 | 0.701 |
| dbmdz/distilbert-base-turkish-cased_finetuned | 0.363 | 0.382 | 0.363 | 0.564 | 0.708 |
| dbmdz/distilbert-base-turkish-cased_MLP | 0.402 | 0.423 | 0.407 | 0.574 | 0.718 |
| savasy/bert-base-turkish-ner-cased_finetuned | 0.418 | 0.436 | 0.412 | **0.654** | 0.755 |
| savasy/bert-base-turkish-ner-cased_MLP | 0.350 | 0.395 | 0.363 | 0.617 | 0.752 |
| mys/electra-base-turkish-cased-ner_finetuned | 0.369 | 0.384 | 0.370 | 0.617 | 0.723 |
| mys/electra-base-turkish-cased-ner_MLP | 0.351 | 0.380 | 0.359 | 0.628 | 0.747 |
| akdeniz27/xlm-roberta-base-turkish-ner_finetuned | **0.500** | **0.505** | **0.497** | 0.638 | **0.792** |
| akdeniz27/xlm-roberta-base-turkish-ner_MLP | 0.438 | 0.471 | 0.450 | 0.617 | 0.775 |

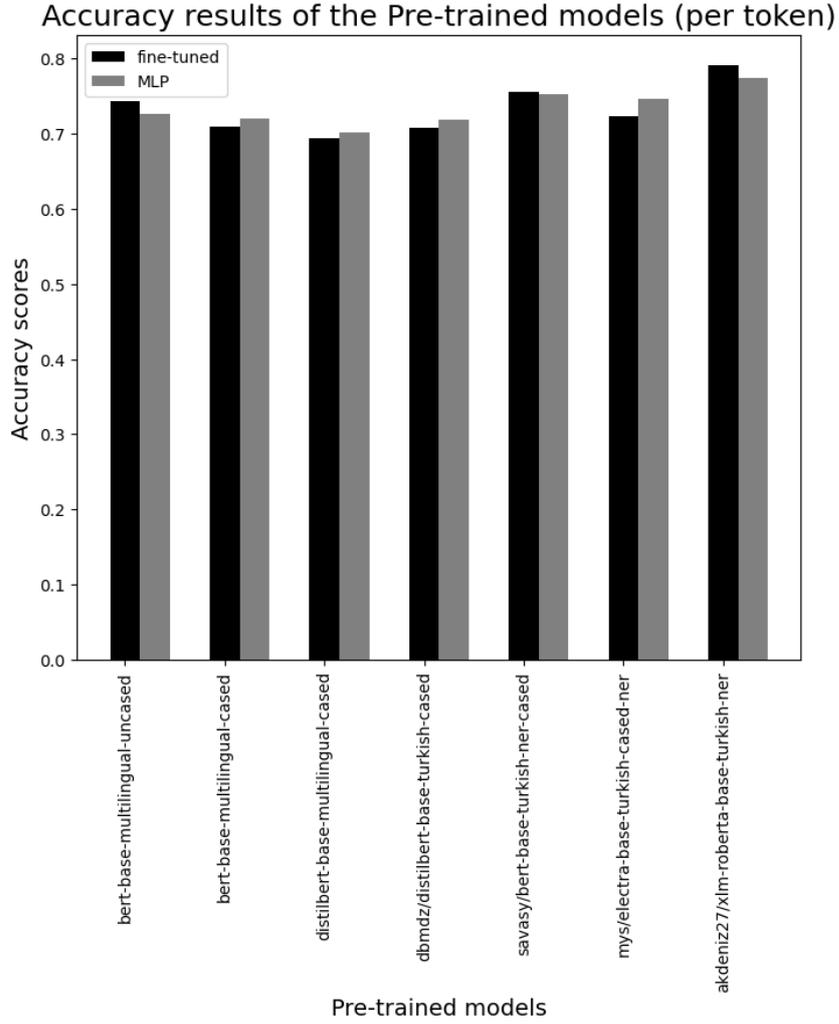

Figure 2. Per token accuracies of fine-tuned and MLP version of the models

**Future Work and Conclusion**

In this paper, we first incorporate Petal Maps address data which is collected from user queries to build a dataset in CoNLL format with a relatively high quality. Then, we utilize the constructed dataset to train models which are decided at the beginning for comparison. We have proposed possible implementation and solutions for Turkish address parsing by using BERT [3] and its variants. Experimental results show that adding additional MLP layer at the top of state-of-art models can slightly improve performance, providing the higher scores for some success metrics. In this experiment;

- Fine-tuning of *savasy/bert-base-turkish-ner-cased* has the highest accuracy for per sample,
- Fine-tuning of *akdeniz27/xlm-roberta-base-turkish-ner* has the highest accuracy for per token and precision (macro), recall (macro) and F1 score (macro).

It can be concluded that language specific models perform better than multilingual models for the map address data. Furthermore the effect of using a two layer MLP for the classification layer can be seen in Figure 2. It appears that a double layer outperforms a single layer in the case of smaller models like ELECTRA and DistilBERT, presumably since the larger models possess a higher representational capacity and can adapt easier to the fine tuning process. Finally as should be expected models fine-tuned on NER tasks yield top performances, demonstrating positive transfer from the NER task to the address parsing task. For future work, models used in the study can be implemented on other large-scale and more complex NLP tasks, which can be reformulated as a sequence labelling task for any other language. Besides which, novel models can be investigated for address data. Moreover our findings with respect to using double layers instead of single layers invite further investigation on other types of tasks. Proving the superiority of the double layer classifier in smaller models would have significant consequences in industrial use cases. Finally seeing that models first fine-tuned in NER tasks perform better at the address parsing task suggests that especially for smaller datasets that gains can be made in the address parsing task by first fine-tuning the model on a large NER dataset. This also merits further investigation.

# References

[1] J. Lafferty, A. McCallum and F. C. Pereira, Conditional random fields: Probabilistic models for segmenting and labeling sequence data., 2001.

[2] S. S. Keerthi and S. Sundararajan, CRF versus SVM-struct for sequence labeling. Yahoo Research Technical Report., 2007.

[3] J. Devlin, M. W. Chang, K. Lee and K. Toutanova, "Bert: Pre-training of deep bidirectional transformers for language understanding.," *arXiv preprint arXiv:1810.04805.,* 2018.

[4] T. Pires, E. Schlinger and D. Garrette, "How multilingual is multilingual BERT?," in *arXiv preprint arXiv:1906.01502*, 2019.

[5] Sanh, V., Debut, L., Chaumond, J., & Wolf, T. (2019), "DistilBERT, a distilled version of BERT: smaller, faster, cheaper and lighter," 2019.

[6] Liu, Y., Ott, M., Goyal, N., Du, J., Joshi, M., Chen, D., ... & Stoyanov, V., "Roberta: A robustly optimized bert pretraining approach," *arXiv preprint arXiv:1907.11692.,* 2019.

[7] A. Virtanen, J. Kanerva, R. L. Ilo, L. J., S. T. J. and S. Pyysalo, "Multilingual is not enough: BERT for Finnish," in *arXiv preprint arXiv:1912.07076.*.

[8] L. Martin, B. Muller, P. J. O. Suárez, Y. Dupont, L. Romary, É. V. de La Clergerie and B. Sagot, "CamemBERT: a tasty French language model.," in *arXiv preprint arXiv:1911.03894.*, 2019.

[9] F. M. Plaza-del-Arco, M. D. Molina-González, L. A. Urena-López and M. T. Martín-Valdivia, "Comparing pre-trained language models for Spanish hate speech detection," *Expert Systems with Applications,* 2021.

[10] A. Balagopalan, B. Eyre, J. Robin, F. Rudzicz and J. Novikova, "Comparing pre-trained and feature-based models for prediction of Alzheimer's disease based on speech.," *Frontiers in aging neuroscience,* 2021.

[11] C. Anand, "Comparison of stock price prediction models using pre-trained neural networks," *Journal of Ubiquitous Computing and Communication Technologies (UCCT),* pp. 122-134, 2021.

[12] Y. Arslan, K. Allix, L. Veiber, C. Lothritz, T. F. Bissyandé, J. Klein and A. Goujon, "A comparison of pre-trained language models for multi-class text classification in the financial domain," in *In Companion Proceedings of the Web Conference*, 2021.

[13] A. Maimaitituoheti, " ABLIMET@ LT-EDI-ACL2022: A RoBERTa based Approach for Homophobia/Transphobia Detection in Social Media," in *In Proceedings of the Second Workshop on Language Technology for Equality, Diversity and Inclusion*, 2022.

[14] A. Özberk and İ. Çiçekli, "Offensive Language Detection in Turkish Tweets with Bert Models," in *2021 6th International Conference on Computer Science and Engineering (UBMK)*.

[15] Z. Guven, "Comparison of BERT Models and Machine Learning Methods for Sentiment Analysis on Turkish Tweets," in *2021 6th International Conference on Computer Science and Engineering (UBMK)*, , 2021.

[16] A. Çelıkten and H. Bulut, "Turkish Medical Text Classification Using BERT," in *2021 29th Signal Processing and Communications Applications Conference (SIU)*, 2021.

[17] G. Z. A., " The Effect of BERT, ELECTRA and ALBERT Language Models on Sentiment Analysis for Turkish Product Reviews," in *6th International Conference on Computer Science and Engineering (UBMK)*, 2021.

[18] A. U. U., B. B. and K. M., "Turkish Sentiment Analysis Using BERT," in *28th Signal Processing and Communications Applications Conference (SIU)*, 2020.

[19] A. Köksal and Ö. Arzucan, "Twitter dataset and evaluation of transformers for Turkish sentiment analysis," in *29th Signal Processing and Communications Applications Conference (SIU)*, 2021.

[20] Clark, K., Luong, M. T., Le, Q. V., & Manning, C. D., "Electra: Pre-training text encoders as discriminators rather than generators," *arXiv preprint arXiv:2003.10555.*, 2020.

[21] T. Akiba, S. Sano, T. Yanase, T. Ohta and M. Koyama, "Optuna: A next-generation hyperparameter optimization framework," in *Proceedings of the 25th ACM SIGKDD international conference on knowledge discovery & data mining*, 2019.

[22] I. Loshchilov and F. Hutter, "Decoupled weight decay regularization.," in *arXiv preprint arXiv:1711.05101*, 2017.

[23] T. Tieleman and G. Hinton, " Lecture 6.5-rmsprop: Divide the gradient by a running average of its recent magnitude. COURSERA: Neural networks for machine learning,," 2012.

[24] H. Robbins and S. Monro, "A stochastic approximation method. The annals of mathematical statistics, 400-407.," 1951.

[25] "Petal Maps," Huawei, [Online]. Available: https://consumer.huawei.com/en/mobileservices/petalmaps/. [Accessed 20 01 2023].

[26] L. A. Ramshaw and M. P. Marcus, "Text chunking using transformation-based learning," *Natural language processing using very large corpora. Springer, Dordrecht.*, pp. 157-176, 1999.

[27] G. Z. A., " Comparison of BERT Models and Machine Learning Methods for Sentiment Analysis on Turkish Tweets," in *6th International Conference on Computer Science and Engineering (UBMK)*, 2021.

[28] Z. Guven, "The Effect of BERT, ELECTRA and ALBERT Language Models on Sentiment Analysis for Turkish Product Reviews," in *6th International Conference on Computer Science and Engineering (UBMK)*, 2021.